\documentclass[runningheads]{llncs}
\usepackage[T1]{fontenc}
\usepackage{graphicx}
\usepackage{rotating}

\usepackage{amsmath, amssymb}
\usepackage{algorithm}
\usepackage{algpseudocode}
\usepackage{tabularray}
\usepackage{hyperref}
\usepackage{subcaption}
\usepackage{tikz} 

\usepackage{adjustbox}

\usepackage[table]{xcolor}

\newcommand{\circnum}[1]{%
  \tikz[baseline=(n.base)]\node[draw=black,fill=gray!15,circle,inner sep=1.6pt,line width=0.5pt](n){\small #1};%
}

\usepackage{multirow}
\usepackage[table]{xcolor}

\usepackage{xcolor}
\definecolor{myPink}{RGB}{255,105,180} 

\begin{document}
\title{Mind the Gaps: A Curated Benchmark for Form Field Detection}
\titlerunning{Mind the Gaps}

\author{
Iheb Brini\inst{1}\orcidID{0000-0003-2326-4669} \and
Omar Moured\inst{2}\orcidID{0000-0003-4227-8417} \and
Hamza Gbada\inst{3}\orcidID{0009-0009-8283-1572} \and
Elisa Barney\inst{1}\orcidID{0000-0003-2039-3844}
}

\authorrunning{I. Brini et al.}

\institute{
Luleå University of Technology, Luleå, Sweden\\
\email{ihebbrini.ing@gmail.com, elisa.barney@ltu.se}
\and
Karlsruhe Institute of Technology, Karlsruhe, Germany\\
\email{omar.moured@kit.edu}
\and
LATIS-Laboratory of Advanced Technology and Intelligent Systems, National Engineering School of Sousse (ENISo), University of Sousse, Tunisia\\
\email{hamza.gbada@eniso.u-sousse.tn}
}
%
%
%
\maketitle              
\begin{abstract}

Form Field Detection (FFD) is a fundamental component of document understanding systems, enabling applications ranging from large-scale industrial digitization to accessible form interaction for automated analysis. Unlike conventional object detection tasks, FFD is inherently challenging because fields are often defined by layout structure and whitespace rather than visible foreground content. Existing large-scale datasets frequently rely on heuristic annotation pipelines, resulting in noisy and inconsistent labels that hinder reliable evaluation.
In this work, we introduce a carefully curated mini-CommonForms benchmark for FFD with consistent, high-quality annotations, and present a detailed evaluation of state-of-the-art detection approaches. The benchmark is designed to support reproducible research and realistic systems in both document automation and accessibility-oriented systems. Dataset and code are available at \url{https://github.com/moured/mini-commonforms}.

\end{abstract}
\section{Introduction}
\label{sec:introduction}

Structured Document Understanding (SDU) seeks to extract structured information from document images to support applications such as industrial document digitization, automated workflow processing, and accessible interaction with forms for visually impaired users. While recent advances in Optical Character Recognition (OCR) and document layout analysis have significantly improved the understanding of textual and semantic regions~\cite{zhong2019publaynet,pfitzmann2022doclaynet}, accurately identifying form input regions remains a challenging and underexplored problem. Reliable Form Field Detection (FFD) is essential for downstream systems including automatic form completion, document automation platforms, and assistive technologies that enable visually impaired users to navigate and interact with structured documents. FFD can also play a major role in document understanding and paired with OCR to locate relevant information from forms. 

\begin{figure}[!h]
\centering
\includegraphics[width=\linewidth]{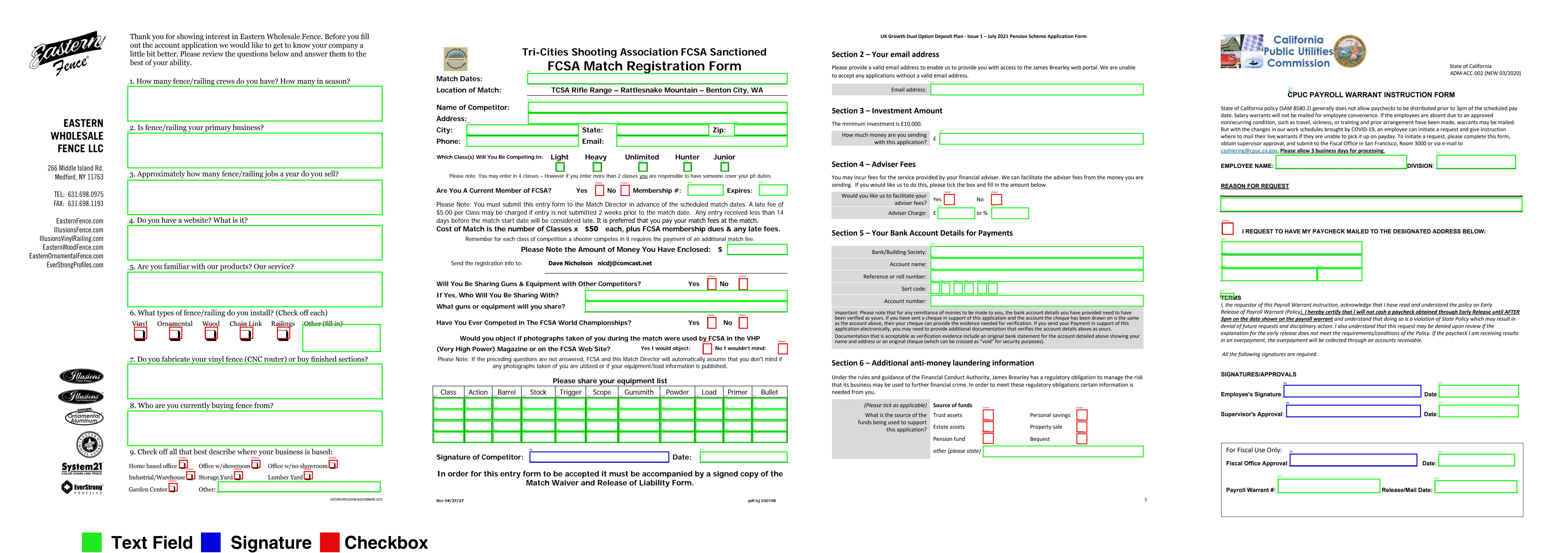}
\caption{Sample annotated documents for the Form Field Detection task \cite{barrow2026commonforms}.}
\label{figure1}
\end{figure}

We define FFD as the task of localizing form input regions such as text boxes, checkboxes, and signature fields in document images. Unlike conventional object detection, FFD is inherently difficult because many targets correspond to empty or weakly defined regions whose structure is expressed primarily through layout organization, whitespace, alignment, or subtle graphical cues rather than visible foreground content. As shown in Figure~\ref{figure1}, valid fields may appear visually ambiguous, loosely bounded, or only implicitly indicated by surrounding context. Consequently, effective FFD requires modeling structural dependencies beyond textual semantics~\cite{appalaraju2021docformer,lee2022formnet}.

The importance of dataset curation has been repeatedly demonstrated in document understanding research. A notable example is FUNSD~\cite{jaume2019funsd}, which transformed the generic RVL-CDIP benchmark~\cite{harley2015icdar} into a foundational resource for structured form analysis. Starting from noisy, low-resolution document scans annotated only with coarse document categories, the authors manually filtered thousands of samples and introduced dense structural annotations, including word-level bounding boxes, semantic labels, and key-value relations. This careful refinement enabled substantial progress in downstream tasks such as layout reasoning and information extraction, illustrating that high-quality annotations can be more impactful than scale alone for structurally complex document tasks.

\begin{figure}[h!]
    \centering
    \includegraphics[width=\textwidth]{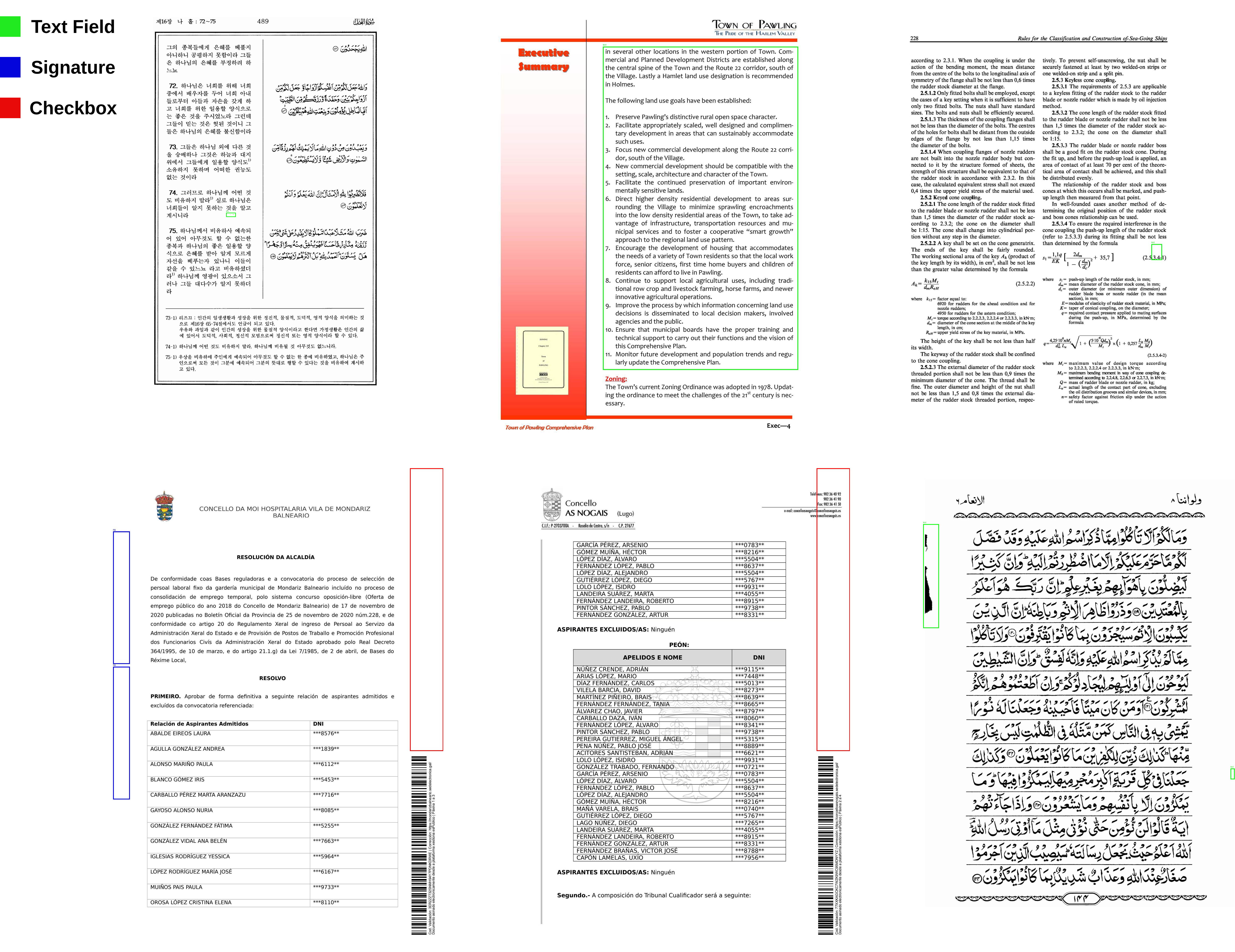}
    \caption{Examples of annotation gaps observed in the CommonForms dataset. The figure illustrates incorrect or imprecise bounding boxes, semantically meaningless or duplicated annotations, missing fields, inconsistent category, multi-template, and challenges arising from multilingual documents and diverse formatting. }
    \label{figure2}
\end{figure}

This observation is particularly relevant for Form Field Detection (FFD), where subtle layout cues and ambiguous field boundaries make annotation quality critical. Although large-scale datasets such as CommonForms~\cite{barrow2026commonforms} provide broad template diversity, heuristic annotation pipelines often introduce substantial noise and inconsistency. As illustrated in Figure~\ref{figure2}, we observe recurring issues including \circnum{1} \emph{geometric noise} (imprecise field localization), \circnum{2} \emph{semantic noise} (missing, duplicated, or ambiguous fields), and \circnum{3} \emph{cross-template inconsistencies} caused by heterogeneous layouts and multilingual formatting conventions. Such artifacts hinder reproducible evaluation and limit the reliability of FFD systems deployed in industrial automation and accessibility-oriented applications.

\begin{enumerate}
\item We introduce \textit{mini-CommonForms}, a carefully curated and human-validated subset of CommonForms, significantly reducing annotation noise and correcting recurring false positive and false negative labeling errors present in the original dataset.
\item We demonstrate that training on the curated benchmark consistently improves Form Field Detection performance, achieving more than 5\% mAP compared to training on the original noisy annotations.
\item We provide the first comprehensive quality analysis of the large-scale CommonForms dataset, identifying recurring annotation issues, structural inconsistencies, and layout-induced failure patterns that impact reliable FFD evaluation and deployment.
\end{enumerate}

\section{Related work}
\label{sec:soa}


\noindent\textbf{Transformer-Based Visual Document Understanding.}
Early works like DocFormer~\cite{appalaraju2021docformer} and DocFormerV2~\cite{appalaraju2024docformerv2} integrated textual, visual, and spatial features through novel self-attention mechanisms to achieve state-of-the-art results on various benchmarks. Similarly, FormNet~\cite{lee2022formnet} and its successor FormNetV2~\cite{lee2023formnetv2} utilized Graph Convolutional Networks (GCNs) and centralized multimodal graph contrastive learning to better capture the structural relationships of tokens in forms. End-to-end approaches such as Dessurt~\cite{davis2022end} and the OCR-free Donut~\cite{kim2022ocr} have further simplified the pipeline by removing the dependency on external OCR engines, generating text autoregressively from document images. While these models excel at general document understanding and key information extraction (KIE), they typically rely on heavy pre-training and often struggle to explicitly model the absence of information, such as empty fields, which are defined by structural gaps rather than textual content.\\


\noindent\textbf{Benchmark Construction and Curated Subsets.}
Creating high-quality benchmarks by curating subsets of larger, noisier collections is a well-established practice in SDU research. For instance, FUNSD~\cite{jaume2019funsd} was created by manually annotating a varied subset of the massive, grayscale RVL-CDIP~\cite{harley2015icdar} collection, providing a challenging testbed for spatial understanding. Similarly, XFUND~\cite{xu2022xfund} extended this concept to multilingual settings, while DocVQA~\cite{mathew2021docvqa} sourced its varied document images from the extensive UCSF Industry Documents \cite{UCSF_IndustryDocumentsLibrary} Library. These datasets demonstrate the value of focused, densely annotated subsets for evaluating specific capabilities like key-value extraction and reasoning. Our work follows this precedent by introducing \textit{mini-CommonForms}, a rigorously curated subset of the CommonForms dataset \cite{barrow2026commonforms}. \\

\noindent\textbf{Domain-Specific Form and Receipt Datasets.}
Beyond subset-based benchmarks, the community has developed numerous datasets targeting specific form-like domains. SROIE~\cite{huang2019icdar} and CORD~\cite{park2019cord} focus on receipt information extraction, with CORD offering a hierarchical schema for complex list structures. For longer, more complex layouts, the Kleister~\cite{stanislawek2021kleister} datasets (NDA and Charity) challenge models to handle multi-page documents. VRDU~\cite{wang2023vrdu} benchmarking suite addresses the "visually rich" aspect with diverse templates and schemas. Other specialized datasets include Form-NLU~\cite{ding2023form} for capturing user intent in form filling and WildReceipt~\cite{sun2021spatial} for processing receipts in unconstrained environments. While valuable, these datasets prioritize text-rich extraction over the structural detection of empty or implicit fields, which is the primary focus of our research. \\

\section{Mini-CommonForms Dataset}
\subsection{Analysis of Common Form dataset}

CommonForms~\cite{barrow2026commonforms} is currently the largest publicly available benchmark for Form Field Detection (FFD). It defines three main field categories as shown in Figure \ref{figure1}: (1) \textit{Text}, which includes regions for written user input; (2) \textit{Choice}, which covers selection elements such as checkboxes and radio buttons; and (3) \textit{Signature}, which refers to areas reserved for signatures, initials, or approval marks. The dataset contains approximately 436k training pages, 33.1k test pages, and 18.2k validation pages, with a total storage size exceeding 163\,GB. Its scale and layout diversity make it an attractive resource for large-scale document understanding research. However, our analysis reveals that the raw dataset contains substantial structural and annotation noise that limits its reliability for rigorous FFD evaluation and training.

First, a significant portion of the dataset consists of low-quality document scans, including blurred pages, compression artefacts, scanning shadows, partially cropped forms, and low-resolution historical documents. Since FFD depends heavily on subtle layout cues such as whitespace, alignment, and faint ruling lines, image degradation disproportionately impacts annotation quality and model learning.

Second, the dataset exhibits strong heterogeneity in document language and formatting conventions. CommonForms contains multilingual documents spanning multiple scripts, layouts, and regional form structures. While this diversity is valuable, it also introduces substantial annotation inconsistencies across templates, especially for structurally ambiguous fields and culturally specific form layouts.

Third, despite heuristic collection procedures, we observe numerous non-form or weakly form-related documents within the dataset. These include letters, invoices, advertisements, and generic scanned pages containing little or no fillable structure. In many cases, annotations are either missing entirely or assigned to visually irrelevant regions, introducing substantial semantic noise during training. We additionally identify several recurrent annotation artefacts, including missing annotations for visually obvious fields, duplicated or overlapping field boxes, invalid bounding boxes with negative or degenerate dimensions, severely misaligned field boundaries, and tiny isolated annotations that likely correspond to parsing artefacts rather than meaningful form fields.

\begin{figure}[h!]
    \centering
    \includegraphics[width=\textwidth]{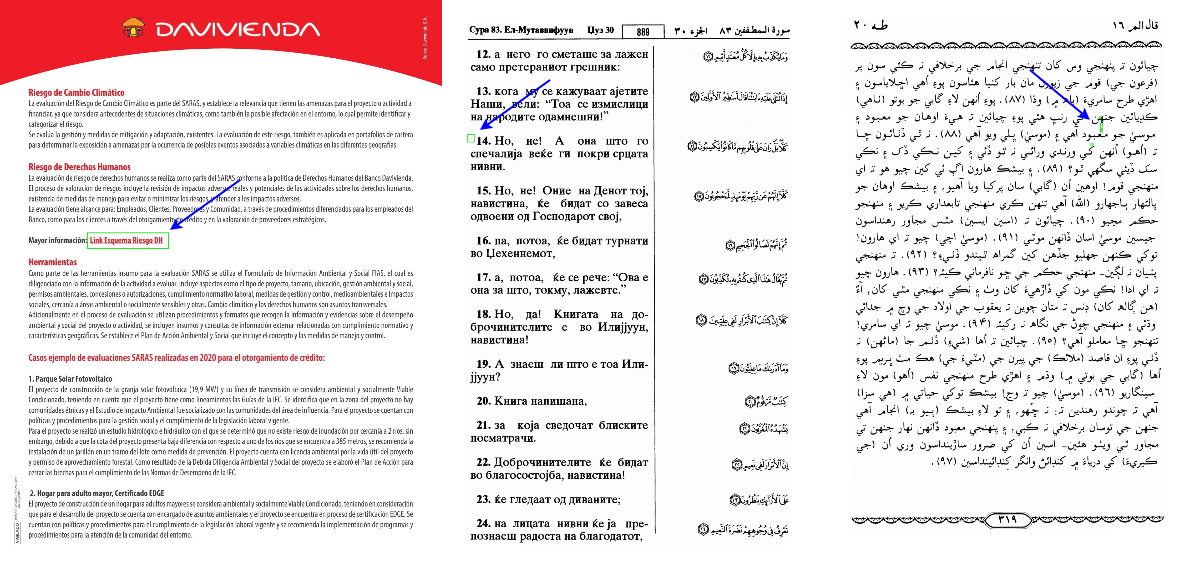}
    \caption{Example of noisy micro-scale bounding box annotations incorrectly assigned within a multilingual non-form document in the CommonForms dataset. Errors such as annotation artefacts and extreme bounding boxes are common throughout the original dataset.}
    \label{fig:bbox_stats}
\end{figure}

Beyond annotation validity, the dataset also exhibits significant statistical irregularities in both bounding box geometry and category distribution. As illustrated in Figure~\ref{fig:bbox_stats}, the bounding box area distribution contains a substantial number of extreme outliers, including numerous annotations with near-imperceptible spatial extent. Many of these regions correspond to isolated marks, OCR artefacts, or annotation noise rather than semantically meaningful form fields. 

Figure~\ref{figure2} illustrates representative examples of these failure cases,  where tiny bounding boxes are incorrectly assigned within a multilingual religious document that does not correspond to a genuine fillable form and can lead to negative behavior during the training process.

\begin{table*}[t]
\centering
\small
\caption{Comparison of related form document datasets.}
\label{tab:dataset_comparison}
\resizebox{\linewidth}{!}{%
\begin{tabular}{l|c|l|l|l|c|c|c|c|c}
\hline
\multicolumn{1}{c|}{\textbf{Dataset}} &
\multicolumn{1}{c|}{\textbf{Year}} &
\multicolumn{1}{c|}{\textbf{Task}} &
\multicolumn{1}{c|}{\textbf{Images}} &
\multicolumn{1}{c|}{\textbf{train/val/test}} &
\multicolumn{1}{c|}{\textbf{FFD}} &
\multicolumn{1}{c|}{\textbf{Empty}} &
\multicolumn{1}{c|}{\textbf{Checkbox}} &
\multicolumn{1}{c|}{\textbf{Signature}} &
\multicolumn{1}{c}{\textbf{Curated}} \\
\hline
\rowcolor{gray!8}
FUNSD \cite{jaume2019funsd} & 2019 & Form Understanding & 199 & 149 / -- / 50 &
$\times$ & $\times$ & $\times$ & $\times$ & $\checkmark$ \\

SROIE \cite{huang2019icdar} & 2019 & Receipt OCR & 987 & 626 / -- / 347 &
$\times$ & $\times$ & $\times$ & $\times$ & $\checkmark$ \\

\rowcolor{gray!8}
AutoFormBench \cite{baral2026autoformbench} & 2026 & FFD & 407 & 70 / 15 / 15\% &
$\checkmark$ & $\checkmark$ & $\checkmark$ & $\times$ & $\checkmark$ \\

CommonForms \cite{barrow2026commonforms} & 2025 & FFD & 486,954 & 435,698 / 18,195 / 33,061 &
$\checkmark$ & $\checkmark$ & $\checkmark$ & $\checkmark$ & $\times$ \\

\rowcolor{gray!8}
Mini-CommonForms (Ours) & 2026 & FFD & 65,736 & 45,285 / 7,487 / 12,964 &
$\checkmark$ & $\checkmark$ & $\checkmark$ & $\checkmark$ & $\checkmark$ \\
\hline
\end{tabular}}
\end{table*}

In addition, the dataset suffers from severe category imbalance. As shown in the category statistics, one annotation class dominates the dataset with more than 225k instances, while another contains fewer than 2k samples. Such imbalance biases model optimization toward majority classes and reduces robustness for structurally important but underrepresented field types.

These observations highlight a central challenge for large-scale Form Field Detection benchmarks: annotation quality and structural consistency are as critical as dataset scale itself. The combination of noisy labels, non-form documents, invalid geometric annotations, and highly imbalanced category distributions can substantially hinder reproducible evaluation and limit the reliability of systems intended for industrial digitization and accessibility-oriented applications.

\subsection{Curation Process}
\subsubsection{Dataset Cleaning and Verification Pipeline}

\begin{figure}[h!]
    \centering
    \includegraphics[width=0.75\linewidth]{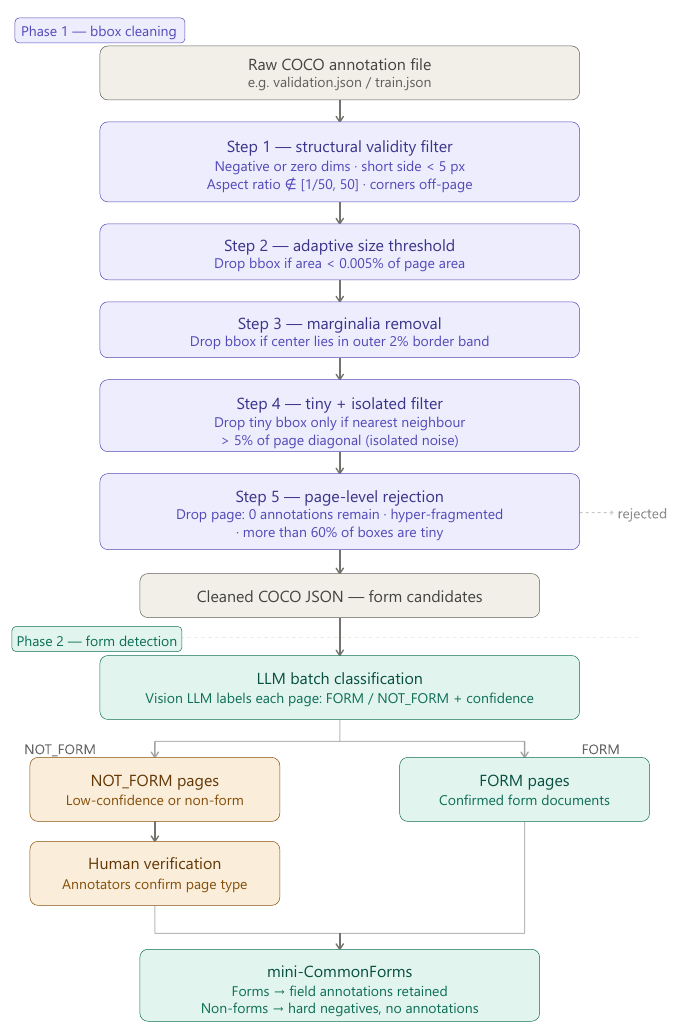}
    \caption{Dataset cleaning and verification pipeline}
    \label{fig:process}
\end{figure}

Based on the original CommonForms COCO annotation, we introduce a multi-stage cleaning and verification pipeline designed to remove annotation artefacts, suppress structural noise, and ensure that retained samples correspond to form documents.
The Figure \ref{fig:process} shows the dataset cleaning and verification pipeline. 

\paragraph{Step 1: Geometric Validation.}
We first remove bounding boxes that violate basic geometric consistency constraints. Specifically, we discard annotations satisfying at least one of the following conditions:

\begin{itemize}
    \item non-positive width or height,
    \item minimum side length smaller than 5 pixels,
    \item extreme aspect ratio outside the interval $[1/50, 50]$,
    \item coordinates extending outside page boundaries.
\end{itemize}

These cases typically correspond to corrupted annotations, parsing artefacts, or degenerate regions.

\paragraph{Step 2: Adaptive Scale Filtering.}
To handle differences in document resolution, we apply a page-relative filtering criterion instead of a fixed pixel threshold. A bounding box is removed if:
\(
\frac{\text{bbox area}}{\text{page area}} < 5 \times 10^{-4}
\)

\paragraph{Step 3: Marginalia Suppression.}
While document borders often contain scanning artefacts, pagination marks, crop remnants, or decorative elements that are not semantically meaningful form fields we remove annotations whose center lies within the outer $2\%$ border band of the page.

\paragraph{Step 4: Context-Aware Tiny Object Filtering.}
Small annotations are not systematically removed, since valid form elements can sometimes occupy very limited space. Instead, a small bounding box is discarded only if it is also spatially isolated. More specifically, an annotation is removed if:

\begin{itemize}
    \item its area is below the adaptive small-object threshold, and
    \item the distance to its nearest neighboring annotation exceeds 5\% of the page diagonal.
\end{itemize}

\paragraph{Step 5: Page-Level Quality Rejection.}
Lastly we evaluate each page globally and discard pages exhibiting severe structural degradation. A page is rejected if at least one of the following conditions holds:

\begin{itemize}
    \item no annotations remain after cleaning,
    \item annotation density is excessively fragmented,
    \item more than $60\%$ of remaining annotations are classified as tiny regions.
\end{itemize}

The impact of this stage is to remove pages dominated by annotation artefacts, fragmented detections, or structurally uninformative layouts.

\subsubsection{Document-Type Verification using Vision-Language Models}

The remaining cleaned pages are treated as candidate form documents. However, large-scale web-collected corpora frequently contain non-form documents despite heuristic filtering. This is related to the issues of wrong annotation as depicted in Figure \ref{fig:bbox_stats}. To address this issue, we introduce a semantic verification stage using a Qwen3.6 \footnote{\url{https://huggingface.co/Qwen/Qwen3.6-27B}} Vision-Language Model (VLM).

The model receives batches of document images and predicts whether each page corresponds to a form document using the following prompt:

\noindent
\colorbox{gray!10}{%
\parbox{0.95\linewidth}{%
\ttfamily\small
You will be shown N images.

For each image, classify whether it is a form, i.e., a document with fields intended to be filled in by a person, such as a tax form, application, intake form, or questionnaire.

\vspace{0.5em}

Respond with exactly N lines, one per image, using the exact filename:

\vspace{0.5em}

\textless filename\textgreater: \textless yes or no\textgreater, confidence \textless 0.0 to 1.0\textgreater
}%
}

\noindent For each image, the VLM produces:
\begin{itemize}
    \item a binary decision (\textit{form} / \textit{non-form}),
    \item a confidence score.
\end{itemize}

The prompt explicitly instructs the model to identify structural indicators of forms, including:
\begin{itemize}
    \item empty text fields,
    \item checkboxes,
    \item signature lines,
    \item fillable tables,
    \item aligned label-field structures.
\end{itemize}

\subsubsection{Human Verification of Negative Samples}

Predictions classified as \textit{non-form} undergo an additional manual Human-in-the loop verification step to reduce false negatives or unknown predictions with low confidence score introduced by the VLM. Documents confirmed as non-form are retained as negative layout examples, while all associated field annotations are removed.

\section{Experiments and Results}
\label{sec:experiments-results}

\subsection{Implementation Details.} 

All detection models were trained with MMDetection~\cite{chen1906mmdetection} on 4$\times$ NVIDIA A100 GPUs for 24 epochs. We used SGD with momentum $0.9$ and weight decay $10^{-4}$, and an initial learning rate of $0.01$ (auto-scaled with base batch size 16). The learning rate employed a linear warm-up for the first 500 iterations (start factor $10^{-3}$), followed by a multi-step schedule with decay factor $0.1$ at epochs 16 and 22. Due to the limited number of images we generated from the LLM calls and the its huge cost, we used a subset of the original training set to train these version of models.
During training, we used aspect-ratio batch sampling and standard augmentations (random resize and horizontal flip). At inference, images were resized to a scale of \textbf{$1333 \times 800$} preserving aspect ratio.

\begin{table}[ht!]
\centering
\small
\setlength{\tabcolsep}{4pt}
\caption{Mini-CommonForms overall and per-class detection performance and evaluation on the original test set from the CommonForms dataset.}
\label{tab:commonforms_map_perclass}
\resizebox{\textwidth}{!}{%
\rowcolors{4}{gray!10}{white}
\begin{tabular}{llccccccccc}
\hline
\multirow{2}{*}{\textbf{Model}} &
\multirow{2}{*}{\textbf{Backbone}} &
\multirow{2}{*}{\textbf{Parameters}} &
\multirow{2}{*}{\textbf{mAP}} &
\multirow{2}{*}{\textbf{mAP$_{50}$}} &
\multicolumn{2}{c}{\textbf{Text}} &
\multicolumn{2}{c}{\textbf{Choice}} &
\multicolumn{2}{c}{\textbf{Signature}} \\
\cline{6-11}
 & & & & &
\textbf{mAP} & \textbf{mAP$_{50}$} &
\textbf{mAP} & \textbf{mAP$_{50}$} &
\textbf{mAP} & \textbf{mAP$_{50}$} \\
\hline
FFDNet-S          & YOLO11              & 9M       & 27.5 & 42.6 & 52.4 & 74.5 & 29.5 & 51.9 & 0.6 & 1.4 \\
FFDNet-L          & YOLO11              & 25M      & 24.1 & 36.9 & 44.6 & 63.9 & 25.2 & 44.0 & 2.5 & 2.8 \\
TOOD              & ResNet-101          & 51.02M   & \textbf{34.6} & 51.5 & 58.9 & 80.3 & \textbf{34.9} & 61.0 & 9.7 & 13.1 \\
DINO              & Swin-L              & 218M     & 33.7 & \textbf{54.2} & \textbf{60.3} & \textbf{83.6} & 34.7 & \textbf{64.3} & 9.3 & 14.6 \\
Def. DETR         & ResNet-50           & 41.075M  & 30.0 & 50.7 & 54.2 & 79.0 & 26.3 & 57.6 & 9.5 & 15.4 \\
Cascade R-CNN     & ResNeXt-101         & 127M     & 28.1 & 43.1 & 46.3 & 70.4 & 22.8 & 44.2 & \textbf{11.9} & \textbf{19.7} \\
Sparse R-CNN      & ResNet-101          & 125.16M  & 26.9 & 41.6 & 46.5 & 64.5 & 24.4 & 45.9 & 9.7 & 14.4 \\
Faster R-CNN      & ResNeXt-101         & 99.651M  & 24.9 & 40.3 & 45.3 & 69.3 & 18.5 & 39.3 & 9.7 & 16.1 \\
GFL               & ResNeXt-101   & 50.88M   & 13.0 & 26.1 & 30.1 & 55.5 & 8.7 & 22.2 & 0.3 & 0.6 \\
\hline
\end{tabular}%
}

\end{table}

\subsection{Evaluation Metrics}
\paragraph{Layout performance.}

We follow the standard COCO-style evaluation protocol for object detection and report mAP$_{50:95}$, together with mAP$_{50}$. In addition to overall scores, we report per-class results for the three Mini-CommonForms categories (\textit{Text}, \textit{Choice}, \textit{Signature}).

\subsection{Results}

\subsubsection{Detection Performance.}
Table~\ref{tab:commonforms_map_perclass} reports the overall and per-class detection performance on Mini-CommonForms. Overall, TOOD achieves the best $\mathrm{mAP}$ (34.6), indicating superior robustness under stricter IoU thresholds (0.50:0.95). In contrast, DINO obtains the highest $\mathrm{mAP}_{50}$ (54.2), suggesting strong coarse localization performance at the IoU threshold of 0.5. At the class level, \textit{Text} detection is the most mature category across models. DINO achieves the best performance for this class (60.3 mAP / 83.6 mAP$_{50}$), followed by TOOD. For the \textit{Choice} category, TOOD achieves the highest mAP (34.9), while DINO obtains the highest mAP$_{50}$ (64.3), showing the advantage of strong global reasoning for structured layout elements such as checkboxes. In contrast, for the \textit{Signature} category, Cascade R-CNN attains the highest mAP (11.9) and mAP$_{50}$ (19.7). This indicates that signature detection remains the most challenging class, likely due to its visual variability and lower representation in the dataset.
From a model-efficiency perspective, Deformable DETR with 41M parameters provides competitive performance compared to heavier CNN-based two-stage models. However, the highest strict-IoU performance (mAP) is achieved by TOOD, highlighting that precise boundary alignment remains challenging for document-layout detection.

\begin{figure}[h!]
    \centering

    \begin{subfigure}{0.31\textwidth}
        \centering
        \includegraphics[width=\linewidth]{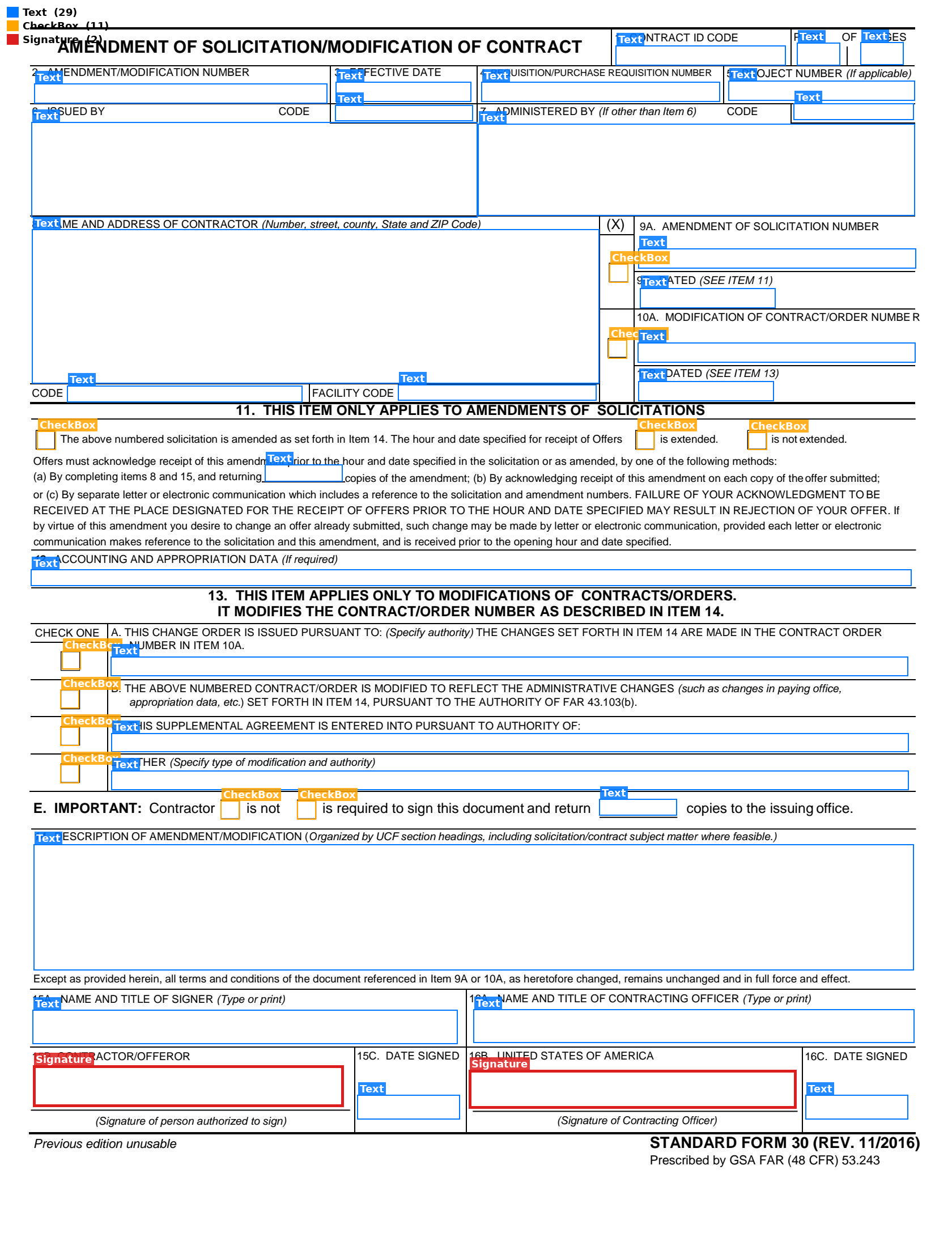}
        \caption{GT}
        \label{fig:qual-gt-1599383}
    \end{subfigure}
    \hfill
    \begin{subfigure}{0.31\textwidth}
        \centering
        \includegraphics[width=\linewidth]{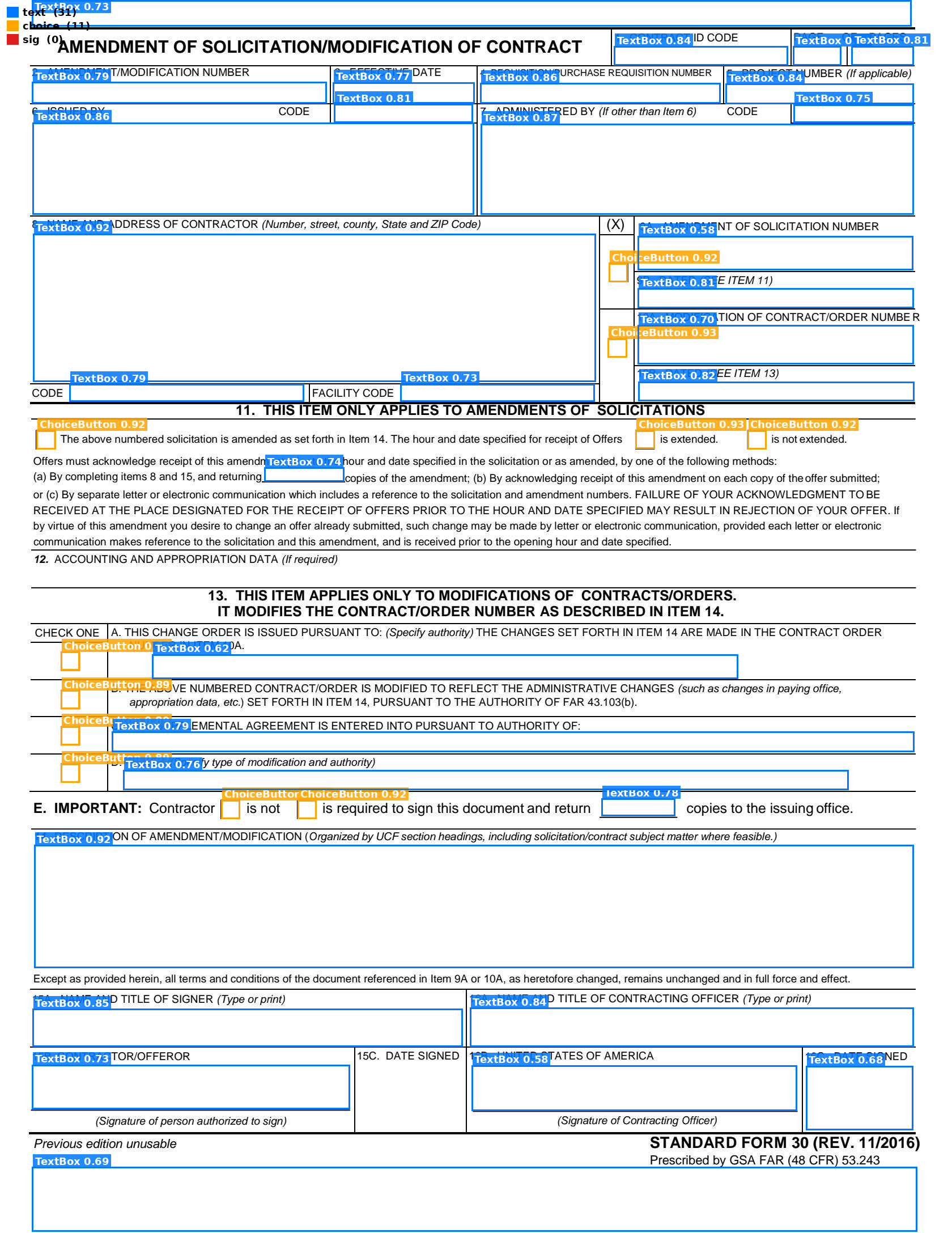}
        \caption{FFDNet-L}
        \label{fig:qual-ffdnet-1599383}
    \end{subfigure}
    \hfill
    \begin{subfigure}{0.31\textwidth}
        \centering
        \includegraphics[width=\linewidth]{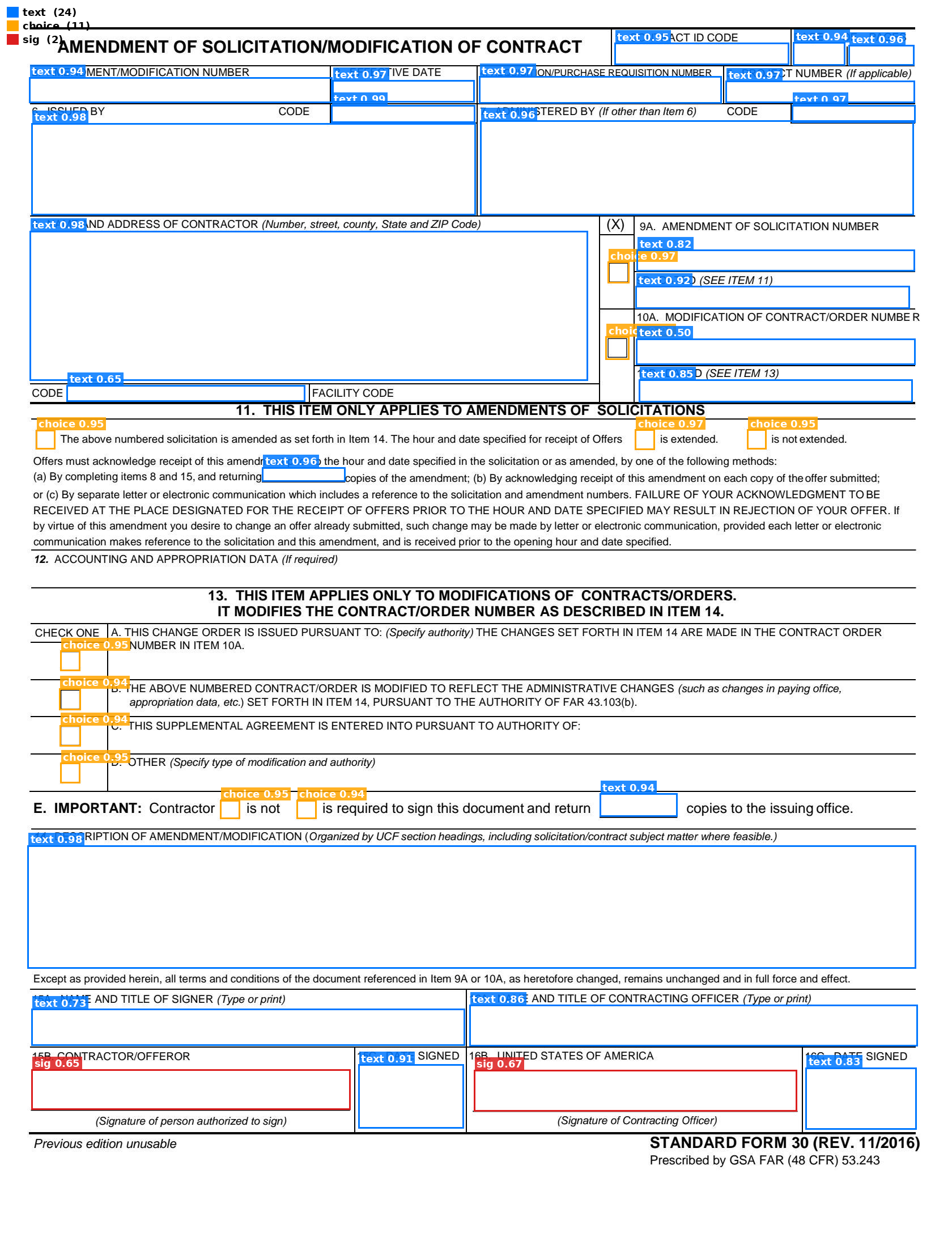}
        \caption{Faster R-CNN}
        \label{fig:qual-fasterrcnn-1599383}
    \end{subfigure}

    \vspace{0.3cm}

    \begin{subfigure}{0.31\textwidth}
        \centering
        \includegraphics[width=\linewidth]{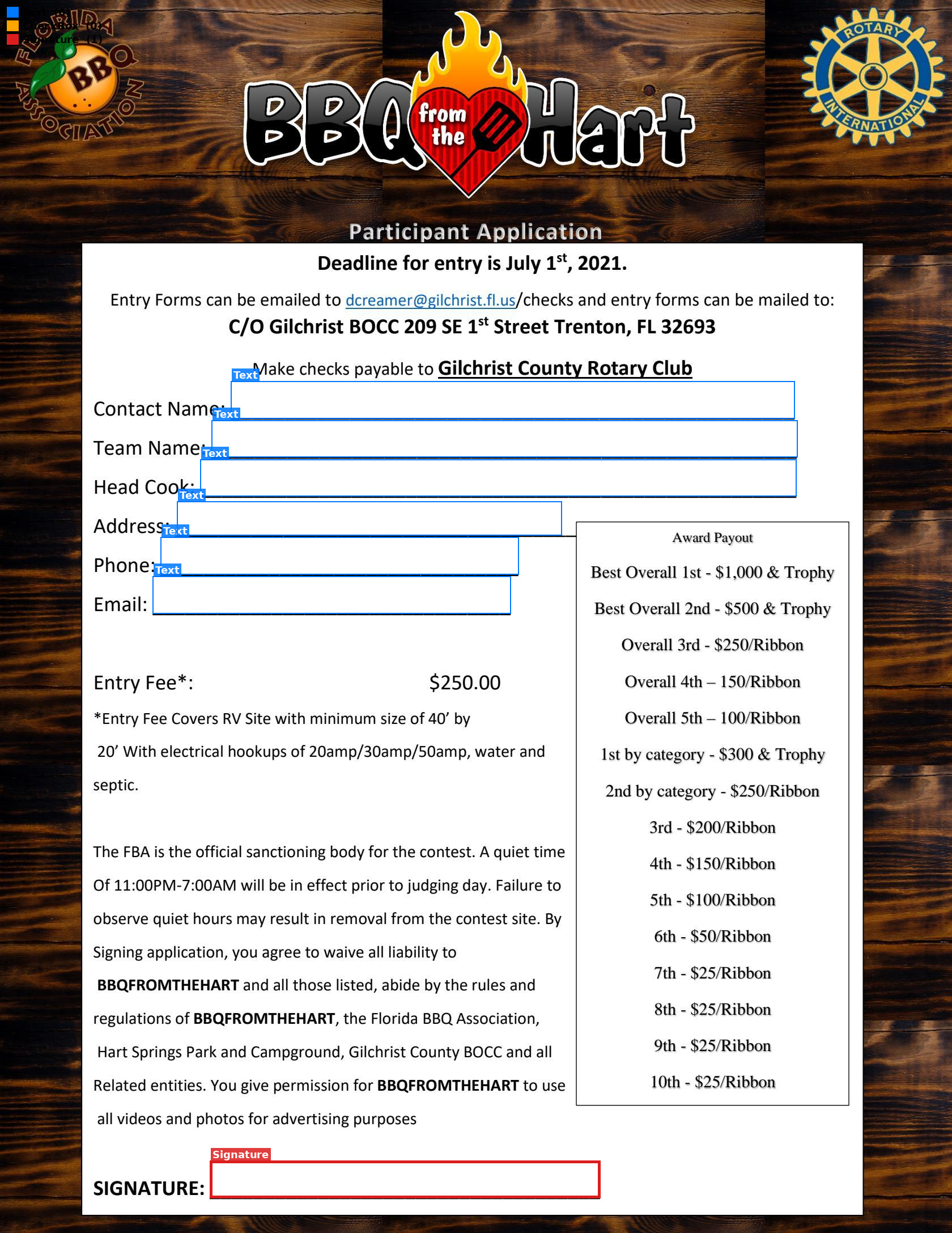}
        \caption{GT}
        \label{fig:qual-gt-6641141}
    \end{subfigure}
    \hfill
    \begin{subfigure}{0.31\textwidth}
        \centering
        \includegraphics[width=\linewidth]{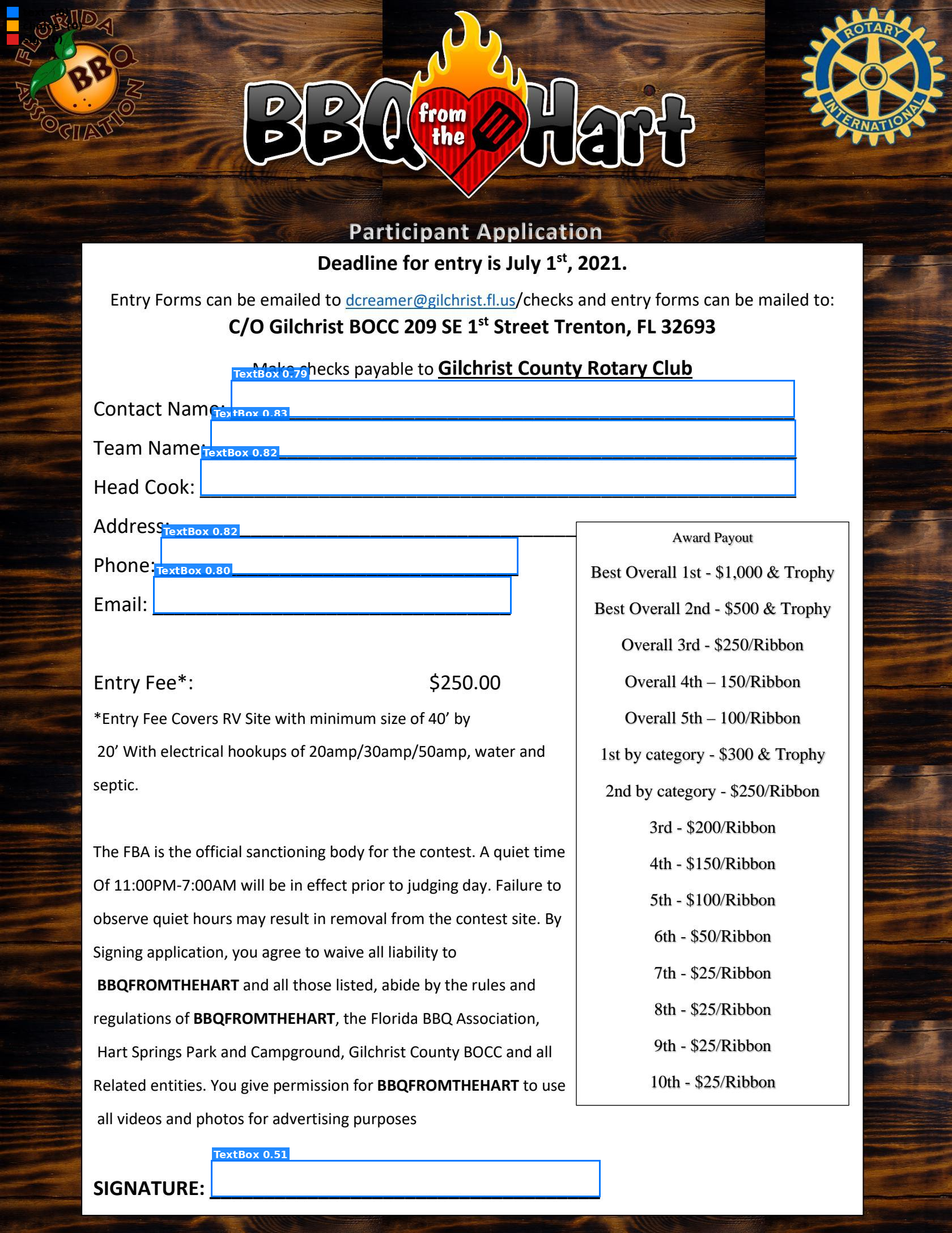}
        \caption{FFDNet-L}
        \label{fig:qual-ffdnet-6641141}
    \end{subfigure}
    \hfill
    \begin{subfigure}{0.31\textwidth}
        \centering
        \includegraphics[width=\linewidth]{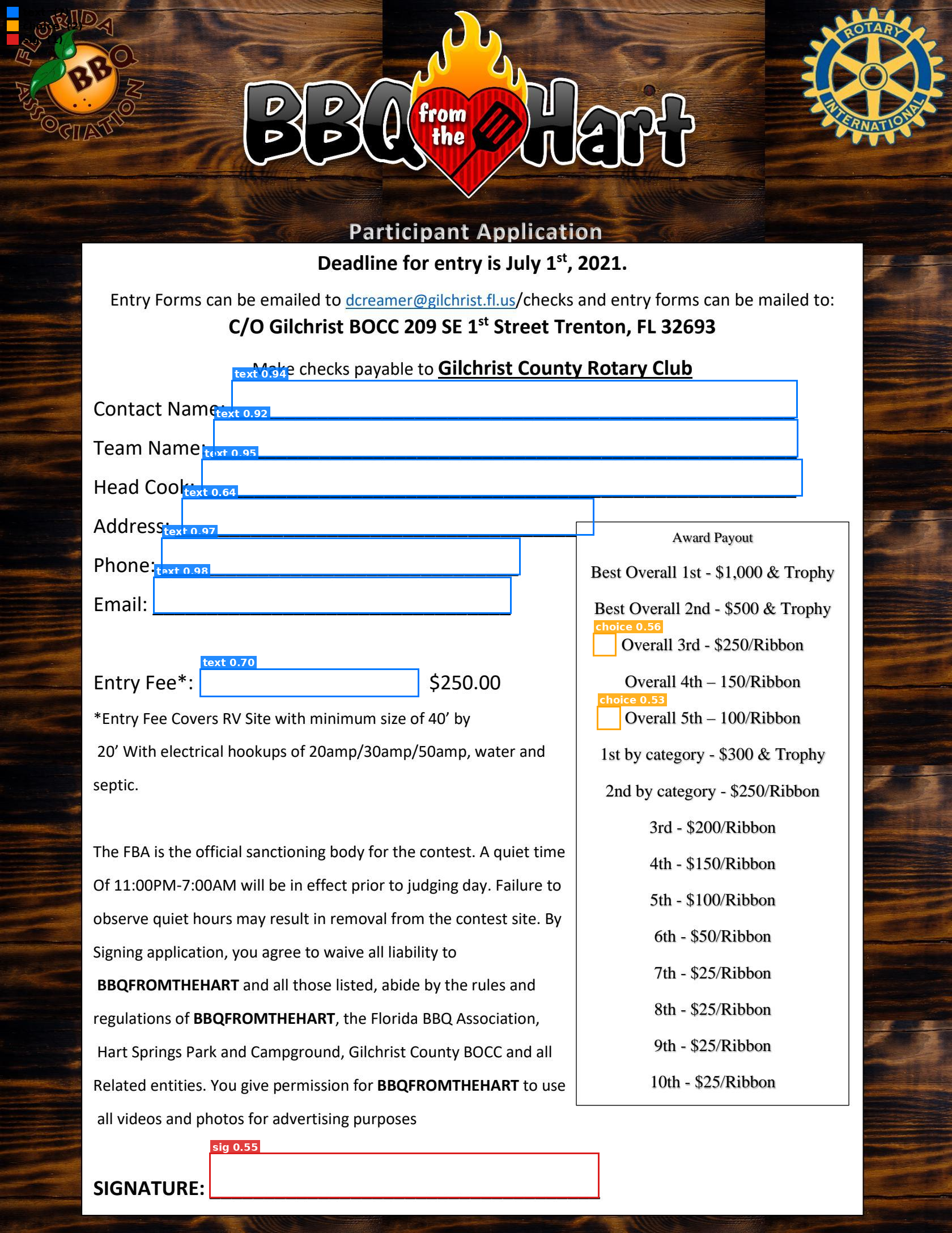}
        \caption{Faster R-CNN}
        \label{fig:qual-fasterrcnn-6641141}
    \end{subfigure}

    \caption{Qualitative comparison between ground-truth annotations, FFDNet-L from the original commonForms paper, and our trained Faster R-CNN model on mini-commonForms samples.}
    \label{fig:qualitative_comparison}
\end{figure}

For FFDNet-S and FFDNet-L, which were trained on the original CommonForms dataset, we observe a noticeable drop in overall mAP, mainly due to the very low performance on the Signature class. Figure~\ref{fig:qualitative_comparison} compares FFDNet-L with our trained Faster R-CNN model. In these examples, signature regions are often predicted as Text fields rather than as Signature fields, which partly explains the degradation in Signature performance. This behavior is also likely related to the strong class imbalance in the original dataset, where the dominant Text category biases the model toward predicting underrepresented classes as Text.

\section{Discussion and Ablation study}
\label{sec:analysis}

As part of our discussion and ablation analysis, we further inspect the behavior of representative detectors. We use two explainability method LRP~\cite{bach2015pixel} and Grad-CAM~\cite{selvaraju2017grad} as qualitative diagnostic tools to examine surface-level failure modes.

\begin{figure}[h!]
    \centering

    \begin{subfigure}{.235\textwidth}
        \centering
        \includegraphics[width=\linewidth, height=4cm]{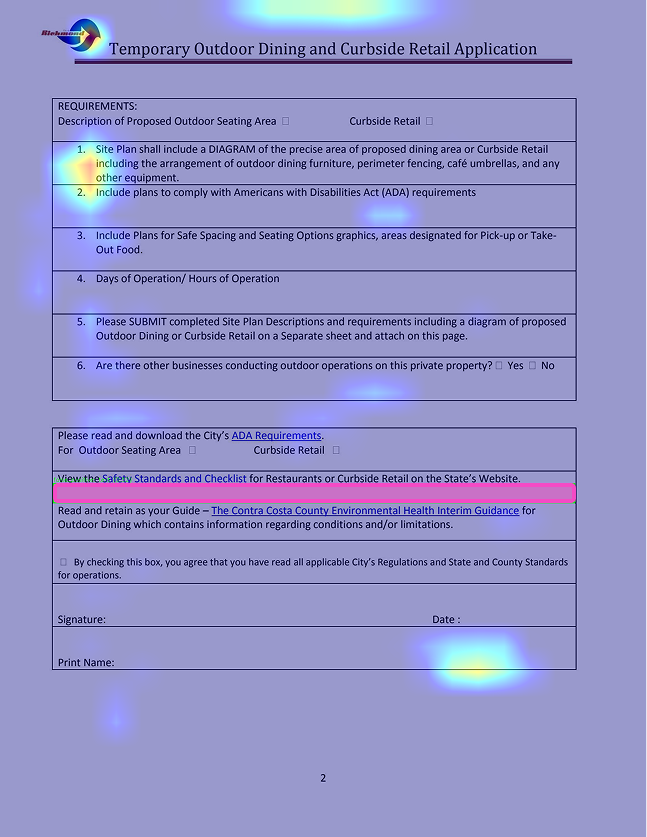}
        \caption{TOOD}
        \label{fig:fail-tood_gradcam}
    \end{subfigure}\hfill
    \begin{subfigure}{.235\textwidth}
        \centering
        \includegraphics[width=\linewidth, height=4cm]{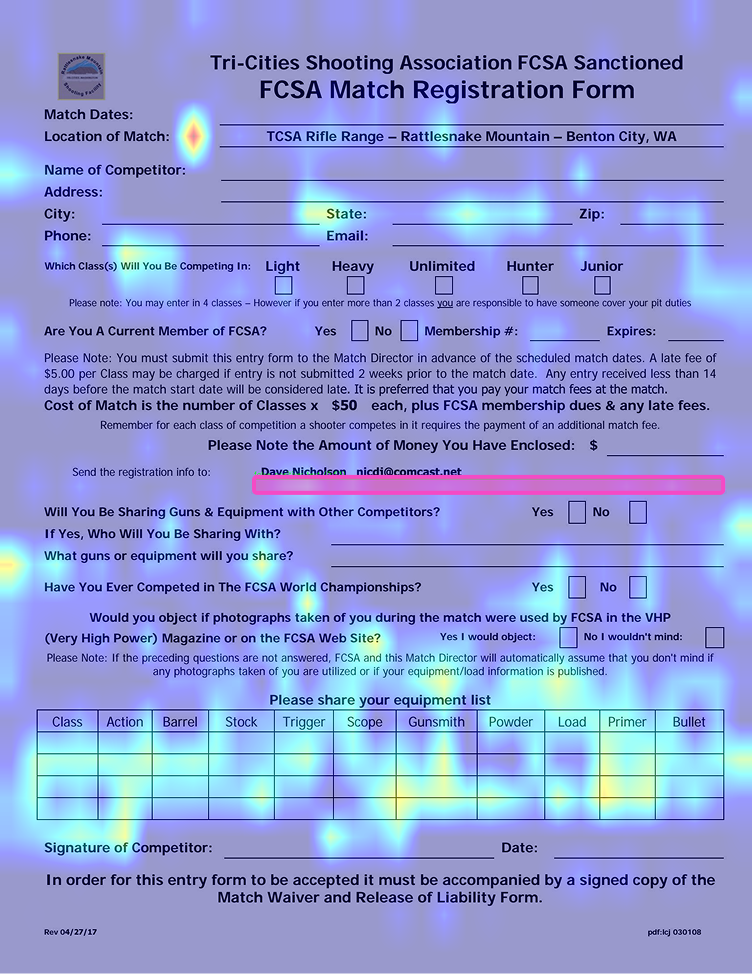}
        \caption{DINO}
        \label{fig:fail-dino_gradcam}
    \end{subfigure}\hfill
    \begin{subfigure}{.235\textwidth}
        \centering
        \includegraphics[width=\linewidth, height=4cm]{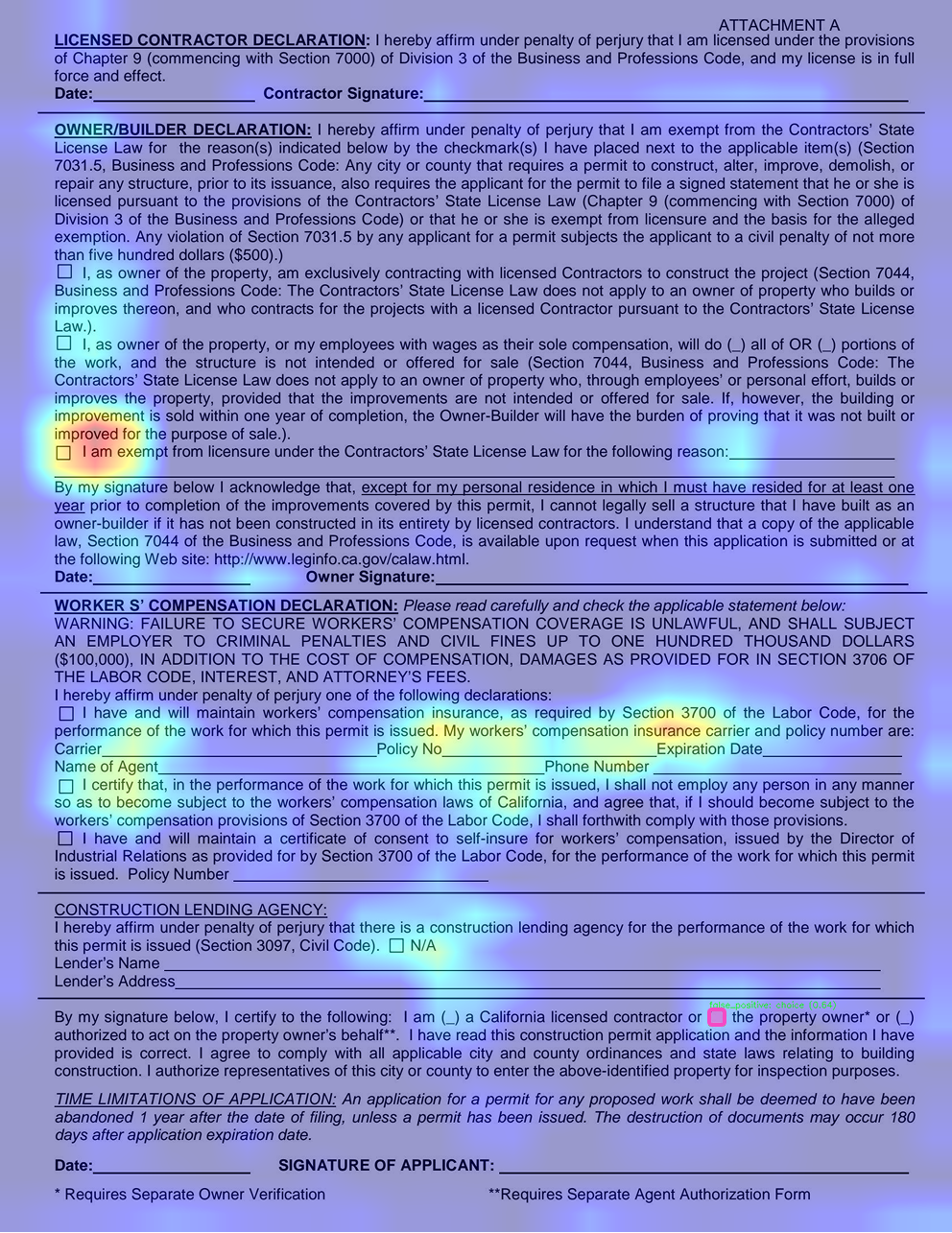}
        \caption{DETR}
        \label{fig:fail-detr_gradcam}
    \end{subfigure}\hfill
    \begin{subfigure}{.235\textwidth}
        \centering
        \includegraphics[width=\linewidth, height=4cm]{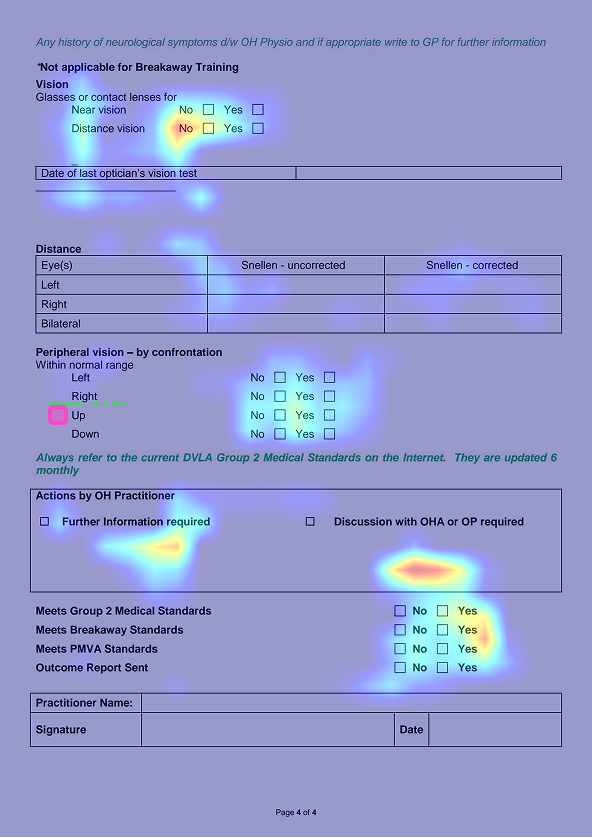}
        \caption{R-CNN}
        \label{fig:fail-rcnn_gradcam}
    \end{subfigure}

    \caption{Representative failure cases in Mini-CommonForms with Grad-CAM. Misclassified boxes are highlighted in \colorbox{myPink}{\strut\textcolor{black}{pink}}.}
    \label{fig:failure-analysis}
\end{figure}

\begin{figure}[h!]
    \centering
        \centering
        \includegraphics[width=0.80\linewidth]{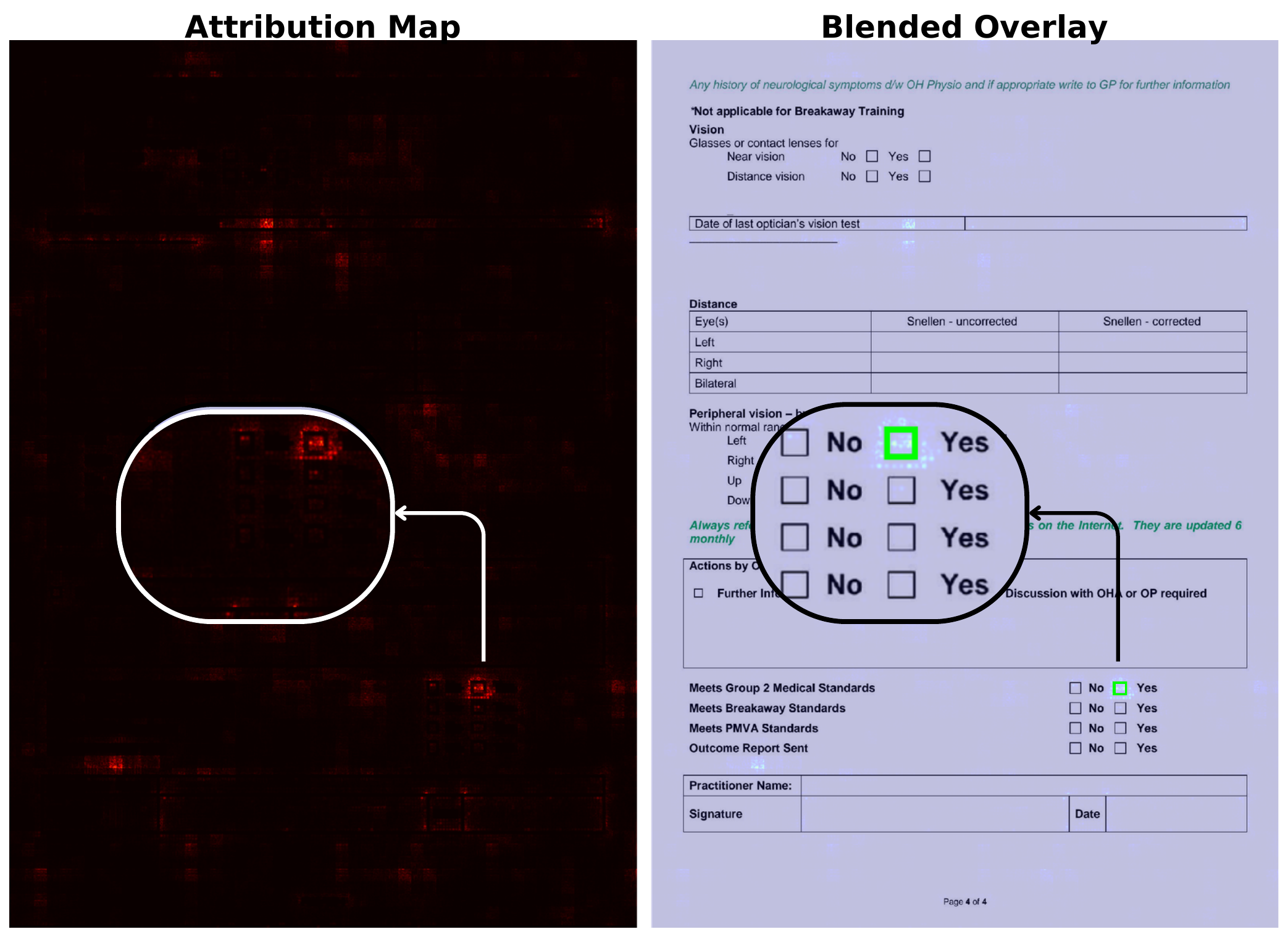}
    \caption{Failure case in the Mini-CommonForm with LRP on Choice target class.}
    \label{fig:lrp-failure_mode}
\end{figure}

Figure~\ref{fig:failure-analysis} presents Grad-CAM explanations for representative failure cases, while Figure~\ref{fig:lrp-failure_mode} shows the corresponding LRP-mixed attributions. In the \textit{Choice} class, Grad-CAM reveals a consistent failure mode across architectures. The detector frequently concentrates on printed label text rather than the checkbox interior, as shown in Figure~\ref{fig:fail-tood_gradcam} and~\ref{fig:fail-dino_gradcam}. The model effectively behaves as a key-field detector, anchoring predictions to semantic tokens instead of localizing the empty input region. This behavior is visible not only in transformer-based models but also in CNN-based Cascade R-CNN, as illustrated in Figure~\ref{fig:fail-rcnn_gradcam}, that indicating a systematic reliance on textual anchors rather than object-internal visual evidence.

For Deformable DETR as depicted in Figure~\ref{fig:fail-detr_gradcam}, attribution often concentrates on the narrow region between label text and predicted box. DINO exhibits a similar pattern, emphasizing relational structures that connect text tokens to nearby box-like regions. In both cases, relevance mass is distributed along spacing and alignment cues rather than within the field interior.

In contrast, LRP-mixed (Figure~\ref{fig:lrp-failure_mode}) highlights whitespace, box contours, and vertical gaps near text. The attribution aligns more closely with structural layout primitives characteristic of form design. Instead of focusing solely on semantic content, LRP captures geometric regularity and empty regions that define input fields.

These qualitative patterns indicate that FFD operates as layout-conditioned reasoning rather than purely object-centric detection.

Unlike natural object detection, form elements are frequently defined by:

\begin{itemize}
    \item Low-texture or empty interiors,
    \item Geometric regularity (rectangles, lines),
    \item Strong spatial dependency on adjacent labels.
\end{itemize}

Correct predictions, therefore, depend on relational cues such as text–field proximity, alignment consistency, and whitespace distribution. Attribution maps reflect this structural dependency: relevance often extends beyond bounding box interiors toward surrounding layout context.

\noindent A consistent architectural trend emerges:
\vspace{3px}

\paragraph{Two-stage CNN detectors} (e.g., Cascade R-CNN) produce more spatially concentrated attributions and higher localization-oriented scores. Region proposal and RoI refinement encourage object-centered evidence aggregation. However, explanations are denser and less stable under perturbation.

\paragraph{Transformer-based detectors} (DINO, Deformable DETR) generate smoother and more robust explanations with lower sensitivity and complexity. Global self-attention distributes relevance across spatial tokens, capturing document-level structure but reducing spatial concentration inside bounding boxes.
\section{Conclusion}
\label{sec:conclusion}

In this work, we introduce mini-CommonForms, a curated compact benchmark derived from CommonForms designed to preserve the characteristics of the original dataset while enabling more reliable evaluation under reduced noise and annotation inconsistencies. 
We benchmark six representative detectors spanning two-stage CNN, one-stage CNN, and Transformer-based architectures, and we analyze their decisions using instance-level post-hoc explanations. Future FFD systems should better exploit document topology, whitespace, alignment, and key-field relationships to improve robustness on complex real-world forms.
\bibliographystyle{splncs04}
\bibliography{references}
\end{document}